# Weakly-Supervised Spatiotemporal Anomaly Detection

Urvi Gianchandani
University of Texas at Dallas

Praveen Tirupattur
University of Central Florida

Dr. Mubarak Shah
University of Central Florida

## Abstract

*In this paper, we explore a weakly supervised method for anomaly detection. Since annotating videos is time-consuming, we only look at weak video-level labels during training. This means that given a video, we know that it is either normal or contains an anomaly, but no further annotations are used to train the network. Features are extracted from video clips that are either normal or anomalous. These features are used to determine anomaly scores for spatiotemporal regions of the clips based on a classifier and the implementation of a multiple instance ranking loss (MIL). We represent both anomalous and normal video clips as positive and negative bags respectively to apply MIL. Furthermore, since anomalies are usually localized to a part of a frame rather than the whole frame, we chose to explore temporal as well as spatial anomaly detection. We show our results on the UCF Crime2Local Dataset, which contains spatiotemporal annotations for a portion of the UCF Crime Dataset.*

## 1. Introduction

The increased use of surveillance cameras in public spaces has led to the problem of a large portion of footage from these cameras going unseen. It is not possible for human monitors to constantly watch surveillance videos to determine when something out of the ordinary is occurring, especially since abnormal events are typically rare. Therefore, it would be more practical to use some type of automated system to recognize and report these kinds of events. This is where the problem of anomaly detection comes into the field of computer vision. Most previous work has been focused on temporal detections of anomalies; in this work, we consider spatiotemporal detection to determine if localizing anomalies within videos would help with the task of anomaly detection.

Anomalies are considered to be abnormal events. Previous work has focused on this particular definition of an anomaly to correctly detect a wide range of different types of anomalies. In this type of approach, normal features are learned during training based on only normal videos. Since normal features are learnt, an abnormal event would be easy to detect, as it is defined as anything that deviates from normal behavior. While this approach is unsupervised and has shown promising results, it is specific to each dataset because normal behaviors are defined by the location of the surveillance cameras.

Recently, work has been done utilizing both normal and anomalous videos during training. Video-level labels and multiple instance learning have been used for anomaly detection [1] Also, the weakly supervised task of anomaly detection using MIL has been mapped to a fully supervised task with noisy labels [3].

In our work, we take inspiration from [1] and use a similar network with some modifications. Instead of extracting features from whole video clips, we divide the feature representations of a video clip into spatiotemporal cubes, such that each region in the feature cube corresponds to a region in the original video clip. Each of these spatiotemporal feature cuboids represents the instances of the positive and negative bags. We use the output from the sixth inception module of the I3D network to extract the features, and the same MIL ranking loss is used to train the network to predict anomaly scores for spatiotemporal regions.

## 2. Related Work

Since anomalies are defined as abnormal events, previous work has been focused on the idea of learning and building a dictionary of normal patterns from normal videos, so that during testing, a normal sample is reconstructed with a low error while an anomalous sample would have a high reconstruction error. This approach has shown promising results and is beneficial, as it does not require annotations for training. A similar approach has also been applied with sparse reconstruction. In this method, a dictionary is built based on sparse representations and a high reconstruction error indicates an anomaly.

The previously mentioned methods all use only one type of videos during training. Recently, work has been done in training using both types of samples [1]. One such approach uses video-level labels to separates clips from normal and anomalous videos into different bags and imposes a multiple instance ranking loss to predict anomaly scores for each video clip. This results in a temporal detection of anomalies within a video since the higher scoring clips in the positive bag would likely contain anomalies.

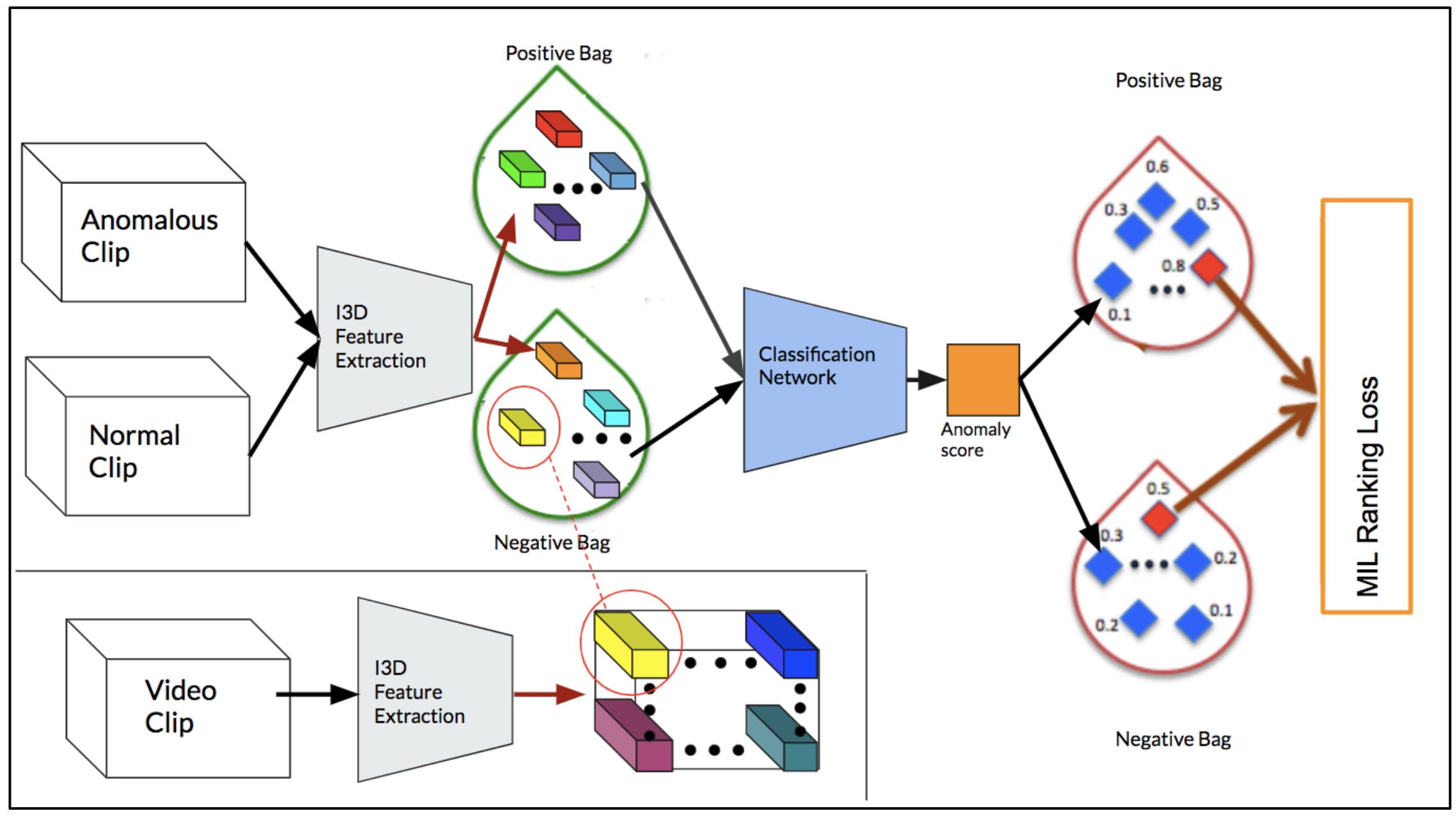

Figure 1: The network architecture of the proposed approach. Given the positive (containing anomaly somewhere) and negative (containing no anomaly) video clips, we extract I3D features and divide each of the feature representations into multiple spatiotemporal cuboids. Then, each video clip is represented as a bag and each spatiotemporal segment represents an instance in the bag. The ranking loss is computed between the highest scores in each bag (show in red in the figure) for training.

Another work defines the weakly supervised anomaly detection problem as a fully supervised action classification task with noisy labels. The labels for normal videos are not noisy while the labels for an anomalous video is noisy because the clips containing the anomaly are unknown. A graph convolutional network is used to clean these noisy labels, and the action classifier and noise cleaner are trained alternately.

Landi *et al.* [2] use a tube extraction method using a set of coordinates, which are the spatial annotations for the anomalies, and a regression network to train the model and detect anomalies. The input tube is used to encode action representation. They use features extracted from the input video as well as from the optical flow using I3D and concatenate them before passing it through an average pooling layer and then the regression network. Since the spatial annotations are used, this is a supervised approach, unlike our method, which is weakly supervised.

## 3. Method

Our approach (Figure 1) is based on the work done by Sultani *et al* [1]. In our approach, videos are divided into segments. We use I3D to extract feature representations of the video clips and we further divided these cuboids into spatial regions, such that each region in the feature representation corresponds to the same region in the original video clip. C3D is used to extract features from video segments in [1] and each segment from either an anomalous or normal video represents an instance of a positive or negative bag. They use the MIL and justify it by stating that given an anomalous video, at least one of the segments will contain an anomaly, whereas none of the segments in a normal video should contain an anomaly. Therefore, they impose the loss such that the maximum anomaly score in the positive bag (anomalous video) is greater than the maximum anomaly score in the negative bag (normal video). In our work, the spatiotemporal cubes make up the instances of the positive (anomalous video segment) and negative (normal video segment) bags. The model is trained using a MIL ranking loss. We use the same logic to say that given an anomalous video clip, at least one of the spatial regions within the clip will contain the anomaly while none of the spatial regions in a normal video clip will have an anomaly.

As mentioned previously, [2] uses a more supervised approach to spatiotemporally detect anomalies. They use what they call 'Tube Extraction' to get a spatiotemporal output from an input video and coordinates. They use I3D to encode features of these tubes, similar to our approach where we encode features from video segments. However, they pass each spatiotemporal tube through I3D whereas we pass a whole video clip through I3D and then divide the feature representations into spatiotemporal cuboids. They use the output from the seventh inception block, while we take the output from the sixth inception block. A two-stream approach is used to combine motion information with the appearance. This appearance and motion volume is passed as input to a regression network to get an anomaly score. Therefore, they take information about both appearance and motion into consideration to train their network, while we focus on appearance. They attempt to train the model with and without the initial coordinates and the results can be seen in Figure 2. The 'Video Segment' column shows their results without the coordinates and the 'Oracle Tube' column shows their results with the coordinates. These results are discussed further later in this paper.

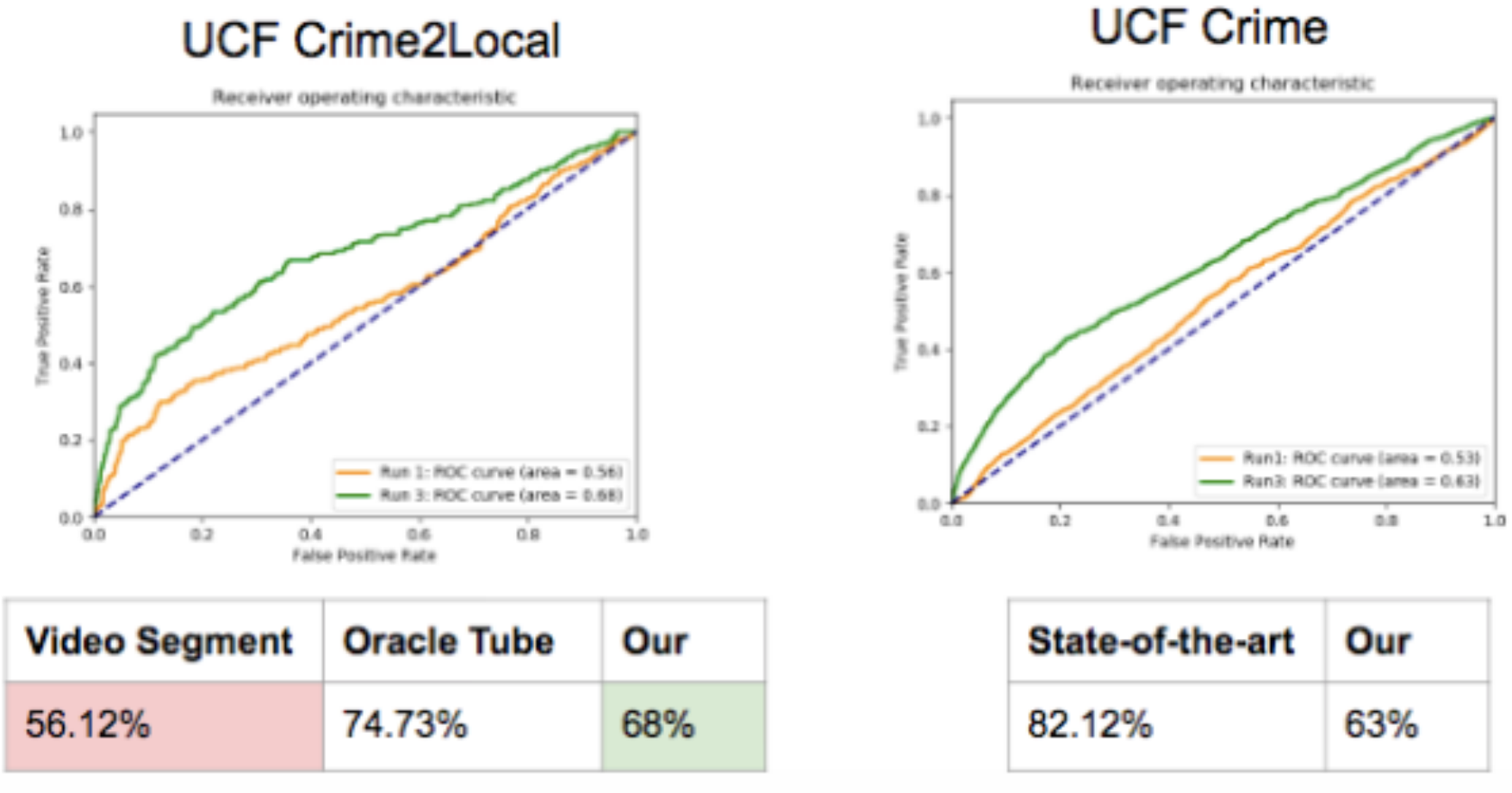

| Video Segment | Oracle Tube | Our |
|---|---|---|
| 56.12% | 74.73% | 68% |

| State-of-the-art | Our |
|---|---|
| 82.12% | 63% |

Figure 2: The graph and table on the left side of the figure are the results on UCF Crime2Local, which contains a portion of the data from the UCF Crime dataset. We compare our best result for AUC of 68%, without using the temporal annotations to the results from [2]. Their weakly-supervised approach gave an AUC of 56.12% whereas their supervised approach gave a result of 74.73%. We compare our results with the 'Video Segment' column since they are both weakly supervised. The graph and table on the right are the results on the whole UCF Crime dataset. We compare our best AUC of 63% to the results from [3].

### 3.1. Multiple Instance Learning

One of the main advantages of using MIL is that it is not necessary to have annotations during training. Video-level labels are known, which means that given a video, we know that it is either normal or anomalous. Therefore, when we divide a video into clips, we are aware of whether the clip is normal or anomalous. As mentioned previously, a positive bag represents an anomalous video clip while a negative bag represents a normal video clip, while the spatial feature representations are the instances of the bags. In a positive bag, it is known that at least one of the instances is anomalous, whereas in a negative bag none of the instances are anomalous. This means that we can apply an objective function based on the maximum scores in each bag.

While pattern detection has been used extensively in the context of anomaly detection, we use a regression-based approach similar to [1]. A ranking loss is imposed such that the maximum anomaly score in the positive bag must be greater than the maximum anomaly score in the negative bag. This is done because the highest score in the positive bag is likely a true positive sample while the highest in the negative bag is likely a false positive sample.

## 4. Experiments

### 4.1. Dataset

Sultani *et al* proposed the UCF-Crime [1] dataset, describing it as a large-scale dataset that represents real-world anomalies. It is made up of both normal and anomalous surveillance videos. For videos with anomalies, there are 13 different types, such as Accident, Fighting, Explosion. The dataset contains 1900 videos with an equal amount of normal and anomalous videos. The training set is made up of 800 normal and 810 anomalous videos while the testing set includes 150 normal and 140 anomalous videos. Since they use a weakly-supervised approach for temporal anomaly detection, there are only temporal annotations for the test set.

Recently, Landi *et al* added spatiotemporal annotations to a portion of the UCF-Crime dataset [2]. This new dataset, called UCF Crime2Local, includes 100 anomalous videos and 200 normal videos, for a total of 300 videos. The training and testing sets contain 210 and 190 videos respectively. Since we are using a weakly supervised approach, we do not use the annotations for training our model.

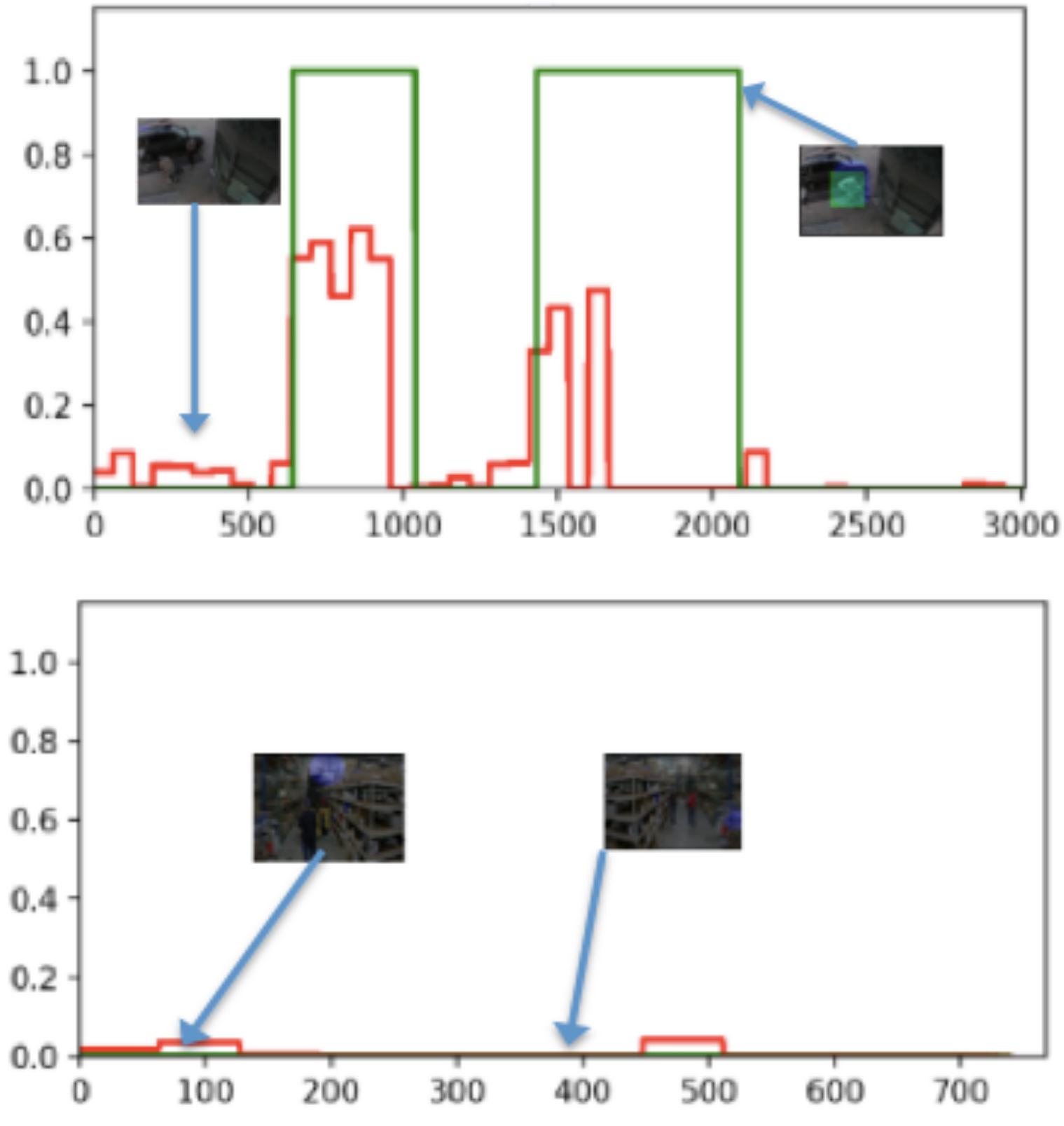

Figure 3: In this figure, we plot the ground truth anomaly score based on temporal annotation. For the frames that contain an anomaly, the ground truth, shown in green, has an anomaly score of one. When the anomaly is not occurring, the ground truth score is zero. The predicted anomaly score from our model is shown in red. It is close to zero when there is no anomaly in the video and increases when the anomaly is occurring. The first graph is from a video that contains an anomaly and the second one is from a normal video.

### 4.2. Implementation Details

Initially, features were extracted from different layers of the I3D to determine which would give the best results. We found that the output from the sixth inception module (4e) has the best results for this problem. A video is divided into segments with 64 frames per segment. Each frame from a video clip is resized to 224 x 224. Each segment is passed through I3D to get a feature representation cuboid. The input to the I3D is [64 × 224 × 224 × 3], where there are 64 frames per clip, each frame is 224 × 224 and has 3 channels (RGB). This is divided along the spatial dimension, resulting in multiple spatiotemporal cuboids for each clip, where each corresponds to a spatial region of the clip. The output of the 4e layer of I3D is of size [528 14 14 4]. This is split spatially into 49 equally sized (2 x 2) cuboids, as shown in Figure 1.

Each of these 49 cuboids is put into the classifier network. The classifier consists of a 3D average pooling layer, followed by five fully connected layers, including batch normalization, ReLU activations, and dropout. Finally, a Sigmoid activation is applied after the final fully connected layer. The output of the classifier is a single score between 0 and 1. Since each of the 49 spatiotemporal feature representations of a clip is passed through the classifier, we get a score for each, which corresponds to an anomaly score for each spatial region of the clip. This process is repeated for a normal clip and an anomalous clip and the multiple instance ranking loss mentioned previously is used to train the model.

### 4.3. Results

To compare our results with other work, we use receiver operating characteristic and its area under the curve (ROC and AUC) to evaluate our model. These results are shown in Figure 2. For the first part, we use only the test set from UCF Crime2Local to determine the area under the curve and compare it with the work done by [2] because they used the same dataset. We compare our weakly supervised method with the ‘Video Segment' and ‘Oracle Tube' results. Since the ‘Video Segment' method they use is not fully supervised, like our approach, we compare our AUC of 68% to their AUC of about 56%. Using a weakly supervised method where no spatiotemporal annotations are known, we get better results. However, they can get an AUC of about 75% with s fully supervised method using ‘Oracle Tubes' where the initial bounding box of the anomaly is known and used during training.

The second part of the figure was computed using the test set of the full UCF Crime dataset so we could compare it to the current state-of-the-art from [3]. Our approach gives us an AUC of 63% whereas they can get about 82% with their method.

Figure 3 shows qualitative results from the experiments. We plot the ground truth anomaly score for the clip (green) and our predicted anomaly score (red). When the ground truth is zero (no anomaly), the predicted anomaly score is relatively low and it increases when the ground truth increases to one.

## 5. Conclusion

In this paper, we explore the impact of considering spatiotemporal features in the task of anomaly detection. We apply the temporal detection based approach in [1] to determine anomaly scores for spatial regions in videos based on feature representation of both normal and anomalous video segments, leading to spatiotemporal detections of anomalies like the work of [2]. We have shown that using our approach to detect anomalies spatiotemporally may be more promising than simply using temporal detections. The method used in this paper may be improved to get results comparable to the current state-of-the-art.

To improve our results, we would need to improve our anomaly detection model. Based on our qualitative results, our model is able to predict the area where an anomaly is occurring in a video, but it often classifies multiple blobs as containing an anomaly. We would need to add some sort of regularization to ensure that only one region is being predicted per video. It would also be beneficial to consider motion information, so using a two-stream approach like [2] including optical flow might improve our network further since motion information has been shown to help in the task of anomaly detection.